\documentclass[11pt]{article}
\usepackage{paclic32}
\usepackage{graphicx}
\usepackage{times}
\usepackage{mathtools}
\usepackage{latexsym}
\usepackage{amsmath}
\usepackage{multirow}
\usepackage{textcomp}
\usepackage{url}

\usepackage{siunitx} 

\title{Word Level Language Identification in English Telugu Code Mixed Data }

\author{Sunil Gundapu \\
  Language Technologies Research Centre \\
  KCIS, IIIT Hyderabad \\
  Telangana, India \\
  {\tt gundapusunil@gmail.com} \\\And
  Radhika Mamidi \\
  Language Technologies Research Centre \\
  KCIS, IIIT Hyderabad \\
  Telangana, India \\
  {\tt radhika.mamidi@iiit.ac.in} \\}

\date{}

\begin{document}
\maketitle
\begin{abstract}
  In a multilingual or sociolingual configuration Intra-sentential Code Switching (ICS) or Code Mixing (CM) is frequently observed nowadays. In the world most of the people know more than one language. The CM usage is especially apparent in social media platforms. Moreover, ICS is particularly significant in the context of technology, health and law where conveying the upcoming developments are difficult in one’s native language.
In applications like dialog systems, machine translation, semantic parsing, shallow parsing, etc. CM and Code Switching pose serious challenges. To do any further advancement in code-mixed data, the necessary step is Language Identification. So, in this paper we present a study of various models - Nave Bayes Classifier, Random Forest Classifier, Conditional Random Field (CRF) and Hidden Markov Model (HMM) for Language Identification in English - Telugu Code Mixed Data. Considering the paucity of resources in code mixed languages, we proposed CRF model and HMM model for word level language identification. Our best performing system is CRF-based with an f1-score of 0.91. 
\end{abstract}

\section{Introduction}

Code switching is characterized as the use of two or more languages, diversity and verbal style by a speaker within a statement, pronouncement or discourse, or between different speakers or situations. This code switching can be classified as inter-sentential (the language switching is done at sentence boundaries) and intra-sentential (the alternation in a single discourse between two languages, where the switching occurs within a sentence). Code mixing is inconsistently elucidated in disparate subfields of linguistics and frequent examination of
phrases, words, inflectional, derivational morphology and syntax use of a term as an equivalent to Code Mixing.

Code mixing is defined as the embedding of linguistic units of one language into utterance of another language. CM is not only used in commonly used spoken form of multilingual setting, but also used in social media websites in the form of comments and replies, posts and especially in chat conversations. Most of the chat conversations are in a formal or semi-formal setting and CM is often used. It commonly takes place in scenarios where a standard formal education is received in a language different than the person’s native language or mother tongue.
 
Most of the case studies state that CM is very popular in the world in the present day, especially in countries like India, with more than 20 official languages. One such language, Telugu is a Dravidian language used by a total of 7.19\% people of Telangana and Andra Pradesh states in India. Because of influence of English, most people use a combination of Telugu and English in conversations. There is lot of research work being carried out in code mixed data in Telugu as well as other Indian languages.

Consider this example sentence from a Telugu website\footnote{www.chaibisket.com} which consists of movie reviews, short stories and articles in cross script code-mixed languages. This example sentence illustrates the mixing being addressed in this paper:

\textbf{Example:} ``John/\textbf{NE} nuvvu/\textbf{TE} exams/\textbf{EN} baaga/\textbf{TE} prepare/\textbf{EN} aithene/\textbf{TE} ,/\textbf{UNIV} first/\textbf{EN} classlo/\textbf{TE} pass/\textbf{EN} avuthav/\textbf{TE} ./\textbf{UNIV}''. (Translation into English: John, you will pass in the first class, even if the exams are well prepared). The words followed by /NE, /TE, /EN and /UNIV correspond to Named Entity, Telugu, English and Universal tags respectively.
In the above example some
words exhibit morpheme level code mixing, like in
\textbf{``classlo''} : \textbf{``class''} (English word) + \textbf{``lo''} (plural morpheme in Telugu). We also consider the clitiques like supere: ’super’ (English root word) + ’e’ (clitique) as code mixed.

We present some approaches to solve the problem of word level language identification in Telugu English Code Mixed data.
The rest of this paper is divided into five sections.

In section 2, we discuss the related work, followed by data set and its annotation in section 3. Section 4 describes the approaches for language identification. And finally section 5 reports Results, Conclusion and Future Work. 

\section{Related Work}

Research on code switching is decades old Gold (1967) and a lot of progress is yet to be made. Braj B., Kachru (1976) explained about the structure of multilingual languages organization and language dependency in linguistic convergence of code mixing in an Indian Perspective. The examination of syntactic properties and sociolinguistic constraints for bilingual code switching data was explored by Sridhar et al. (1980) who contended that how ICS
shows impact on bilingual processing. According to Noor Al-Qaysi, Mostafa Al-Emran (2017) code switching in social networking websites like Facebook, Twitter and WhatsApp is very high. The result of such case studies indicated that 86.40\% of the students using code switching on social networks, whereas 81\% of the educators do so.

One of the basic tasks in text processing is Part of Speech (POS) Tagging and it is primary step in most of NLP. Word level Language Identification can be looked at as a task similar to POS tagging. For POS tagging task Nisheeth Joshi et al. (2013) developed a Hidden Markov Model based tagger in which they take one word in a sentence as a data point. To figure out this sequence labeling classification problem they consider the words as observations and POS tags as the hidden states. Kovida at el. (2018), with the use of internal structure and context of word perform POS tagging for Telugu-English Code Mixed data.

Language Identification issue has been addressed for English and Hindi CM data. For this problem Harsh at el. (2014) constructed a new model with the help of character level n-grams of a word and POS tag of adjoining words. Amitava Das and Bjrn Gambck (2014) modeled various approaches like unsupervised dictionary based approach and supervised SVM word level classification by considering contextual clues, lexical borrowings, and phonetic typing for LI in code mixed Indian social media text.

Ben King and Steven Abney (2013) elaborate the LI problem as a monolingual text of sequence labeling problem. The authors used a weakly supervised method for recognizing the language of words in the mixed language documents. For language identification in Nepali - English and Spanish - English data Utsab Barman et al. (2014) extract the features like word contextual clues, capitalization, char N-Grams and length of word and more features for each word. These features are input for the traditional supervised algorithms like K-NN and SVM.

Indian languages are relatively resource poor languages. Siva Reddy and Serge Sharoff (2011) are modeled a cross language POS Tagger for Indian languages. For an experimental purpose they used the Kannada language by using the Telugu resources because Kannada is more similar to Telugu. To do effective text analysis in Hindi English Code mixed social media data Sharma et al. (2016) are stimulate the shallow parsing pipeline. In this work the authors modeled a language identifier, a normalizer, a POS tagger and a shallow parser.

Most of the experiments have been carried out by using dictionaries, supervised classification models and Markov models for language identification. Finally, most of the authors concluded that language modeling techniques are more robust than dictionary based models.

We attempted this language identification problem with different classification algorithms to analyze the results. According to our knowledge, until now no work has been done on language identification in Telugu-English Code-mixed data.

\section{Dataset Annotation and Statistics}

We use the English Telugu code mixed dataset for language identification from the Twelfth International Conference on Natural Language Processing (ICON-2015). The dataset is comprised of Facebook posts, comments, replies to comments, tweets and WhatsApp chat conversations. This dataset contains 1987 code-mixed sentences. These sentences are tokenized into words. And the tokenized words of each sentence are separated by new line.
The dataset contains 29503 tokens.

\begin{table}[h!]
\begin{center}
\begin{tabular}{|c|c|c|}
\hline 
\bf Language  & \bf Label & \bf Percentage\\
\bf Label  & \bf Frequency & \bf of Label \\
\hline
Telugu & 8828 & 29.92 \\ 
English & 8886 & 30.11 \\ 
Universal & 11033 & 37.39 \\ 
Named Entity & 756 & 2.56 \\
\hline
\end{tabular}
\end{center}
\caption{\label{font-table}  Statistics of Corpus. }
\end{table}

Each word is manually annotated with its POS tag and language identification tag which are: Telugu (TE), English (EN), Named Entity (NE) and Universal (UNIV). Out of the total data, 20\% data is kept aside for testing and the remaining data is used for
training the model. 

\section{Approaches for Word Level Language Identification (LI) }

LI is the process of assigning a language identification label to each word in a sentence, based on both its syntax as well as its context. We have implemented baseline models using Nave Bayes Classifier and Random Forest Classifier with term frequency
(TF) representation. We also performed experiments using Hidden Markov Model (HMM) with Viterbi Algorithm and Conditional Random Field (CRF) using python crfsuite as described in the following sections. 

\subsection{Naïve Bayes Classifier using TF-IDF Vectors}

The baseline Nave Bayes model implemented for language identification in English Telugu Code Mixed data produced estimable results. We are considered this LI problem similar to a Text Classification problem. We consider each sentence as a document and each word in it as a term, for which we calculate the conditional probability for each class label. The Nave Bayes assumption is that independence among the features, each feature is independent to the other features. We first convert the input code mixed words into TF(Term Frequency) - IDF(Inverse Document Frequency) vectors.

\textbf{Word to Vector:} Initially, the raw text data will be processed into feature vectors using the existing dataset. We used the TF-IDF Vector as the features for both the trigrams and character trigrams. TF-IDF calculates a score that represent the importance of a word in a document.
 
\begin{itemize}
 
\item TF: Number of times term word(w) appears in a document / Total no. of terms in the document
\item Inverse Document Frequency (IDF): loge (Total no. of documents /
No. of with term w in it)
\end{itemize}

We trained the model using set of \textbf{``n''} word vectors and corresponding labels.

\begin{itemize}
\item Input Data: Word vector of each word in corpus \[(w_1\_vec, w_2\_vec,....., w_n\_vec) \]
\item Class Labels: (EN, TE, UNIV, NE)
\item Model: \textbf{(Input word vectors)} $\,\to\,$ \textbf{(Labels)}
\end{itemize}

The baseline Nave Bayes model with character level TFIDF word vectors performed with an accuracy of 77.37\% and 72.73\% with trigram TFIDF word vectors.

\subsection{Random Forest Classifier with Count Vectors}
Random forest is an ensemble classifier which selects the subset of random code mixed observations and subset of class labels or variables from training data. For each subset of sample of training observations it sets up a decision tree after extracting the features. All these decision trees are consolidated together to get the prediction using a voting procedure.

\begin{figure}[h!]
  \includegraphics[width=\linewidth]{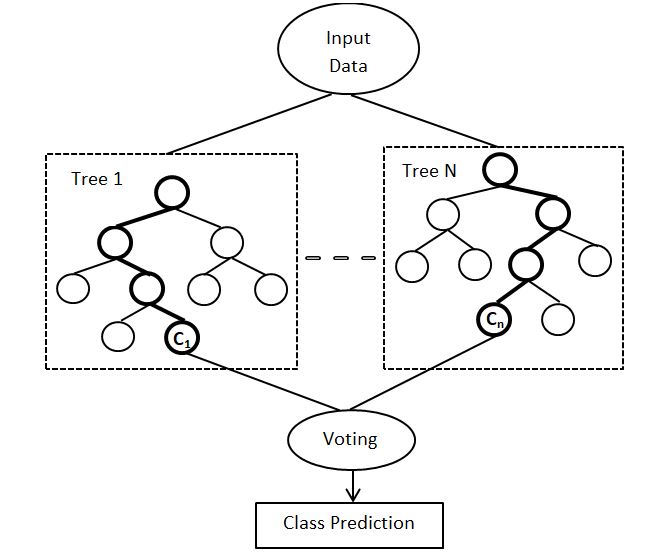}
  \caption{Random forest classifier model.}
  \label{fig:boat1}
\end{figure}

Initially, the whole observations in code mixed dataset are converted into vector form (Count Vector) for feature extraction. In a count matrix, a column represents a term from the dataset, a row represents the current word and the cell contains the frequency of the current word in the corpus. 

We used 50 decision trees in the random forest for our experiments.

Random Forest automatically calculates the relevance score for each feature in training phase. The summation score of each feature is normalized to 1. These scores assist the model to select the significant features for model labeling. We use the \textbf{Mean Decrease in Impurity (MDI)} or \textbf{Gini Importance} to calculate the importance of each feature. 

For this English Telugu Code Mixed corpus it performs with an accuracy of 77.34\%, but since the dataset had very few sentences (1983 sentences, 31421 words) to construct the decision trees, we suspect overfitting was an issue.

\subsection{Hidden Markov Model}

HMM inherently takes sequence into account. We formulate the problem such that an entire sentence - sequence of words is a single data point. The observations are the words and the hidden states are language tags. This HMM based language tagging
assigns the best tag to a word by calculating the forward and backward probabilities of tags along with the sequence provided as an input.

\[P(tag_i|word_i)\]
\[ \Updownarrow \]
\[ P(tag_i|tag_{i+1})*P(tag_{i+1}|tag_i)*P(word_i |tag)\]

In above equation, \(P(tag_i|tag_{i-1})\) probability for current tag given by the previous tag and \(P(tag_{i+1}|tag_i)\) is the probability of the future tag given the current tag for focused word. Here the \(P(tag_{i+1}|tag_i) \)indirectly gives the meaning of transition of two tags.

\begin{figure}[h!]
  \includegraphics[width=\linewidth]{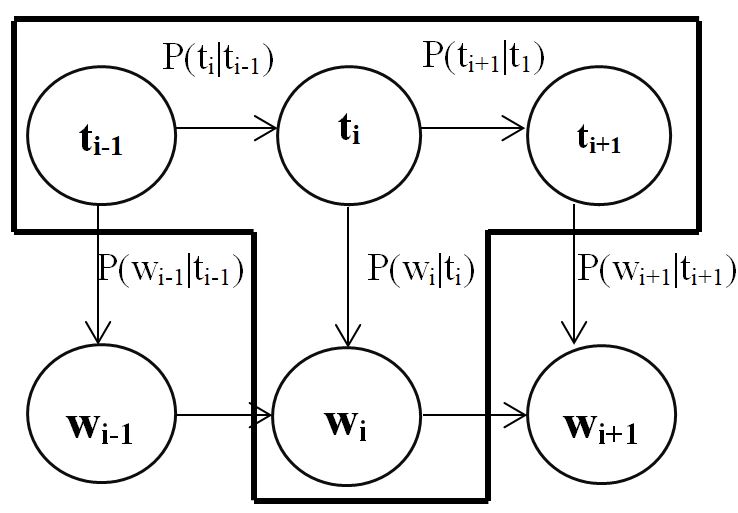}
  \caption{Context Dependency of HMM.}
  \label{fig:boat1}
\end{figure}

For each tag transition probability is computed by calculating the frequency count of two tags seen together in the corpus divided by the frequency count of the previous tag seen independently in the
training corpus.

The likelihood probabilities calculated using \(P(word_i|tag_i)\) i.e. the probability of the word given a current tag. This probability is computed using

\[P(word_i|tag_i)\]
\[ \Updownarrow  \]
\[ Freq(tag_i, word_i)/Freq(tag_i) \]

Here, the probability of word provided a tag is computed by calculating the frequency count of the tag in the sentence and the word occurring together in the corpus divided by the frequency count of the occurrence of the tag alone in the corpus.

For testing the performance of the model, the corpus was divided into two parts: 80\% for training, 20\% for testing. The model performs with an accuracy of 85.46\%. Perhaps, with more features, the accuracy could be further improved.

\subsection{Conditional Random Field}

Conditional Random Field (CRF) has been implemented for word - level language identification with the help of crf pysuite. CRF which is a simple, customizable and open source implementation. The most significant part of this approach is that feature selection. We proposed various feature set templates based on different possible combinations of available words, it's context and possible tags.

\begin{figure}[h!]
  \includegraphics[width=\linewidth]{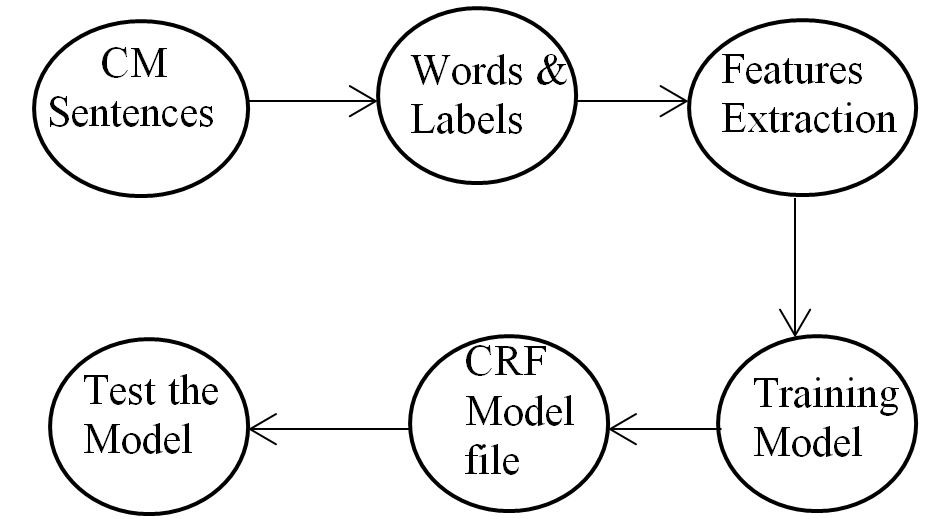}
  \caption{Steps involved in CRF Model.}
  \label{fig:boat1}
\end{figure}

For this model, we are use the following feature set to train the CRF Model.
\begin{itemize}
\item \textbf{Current word and its POS tag:} We consider the current testing word and its POS tag to predict the language label.  
\item \textbf{Next word and its POS tag:} We define the next word and its POS tag to capture the relation between current and next word. 
\item \textbf{Previous word and its POS Tag:} To extract the context of current word, we took the previous word and its POS tag based on first order markov assumption.  
\item \textbf{Prefix and Suffix of focus word:} Extracted the
prefix and suffix of current word. If the word doesn’t have any prefix or suffix we add NULL as a feature.
\item \textbf{Length of word:} We considered the length of the word as one of the feature.   
\item \textbf{Start with numeric digit:} Whether the word
starts with a numeric digit or not eg. \textbf{``2morrow''} (tomorrow).  
\item \textbf{Contains numeric digit:} Whether the word
contains any number eg. \textbf{``ni8''} (night). A regular expression was used to detect this feature.  
\item \textbf{Start with special symbol:} Whether the word
starts with any special symbol or character like !,@,\$,/...etc.
\item \textbf{Start with capital letter:} This feature is TRUE if the word starts with capital letter else FALSE. 
\item \textbf{Contains any capital letter:} Whether the
word contains any capital letter like \textbf{``aLwAyS''} (always).
\item \textbf{Previous word language tag:}  To predict the
current word language label we considered the previous word language label.
\item \textbf{Character N-grams (Uni-, Bi-, Trigram of the word) :} Lot of code mixed words written in different formats. Example like (\textbf{akkada} → \textbf{ekkada, yekkada, aeikkada}). To obtain the syntactical information of the word we took the uni-, bi-, trigram of the word in forward and backward direction into account. The uni-, bi-, trigrams of word in forward (\textbf{a, ak, akk}), backward (\textbf{a, da, ada}) of word respectively.
\end{itemize}

Above features are taken into account to tag the language label for current word. While testing, we assign untagged words such as URLs containing http:// or .in or www or smileys with a default tag of UNIV (universal tag). 

\begin{table}[h!]
\begin{center}
\begin{tabular}{|c|c|c|c|}
\hline 
\textbf{Label}  &  \textbf{Precision} &  \textbf{Recall} &  \textbf{F1- Score} \\ \hline
Telugu & 0.84 & 0.81 & 0.87 \\
English & 0.88 & 0.87 & 0.88 \\
NE & 0.93 & 0.93 & 0.95 \\
Universal & 0.48 & 0.39 & 0.54 \\ \hline
Average & 0.89 & 0.90 & 0.91 \\
\hline
\end{tabular}
\end{center}
\caption{\label{font-table}  Experimental results of each tag. }
\end{table}

With the help of scikit learn, Natural Language Took Kit and python crfsuite we performed the experiments on Engish Telugu corpus. The train corpus contains 23,635 words and test corpus contains 5868 words. We applied three-fold cross validation on our corpus for all experiments. The above feature set gave the highest accuracy of 91.2897\%.
\section{Results and Observations}
The language identification was performed done by Naive Bayes and Random Forest classifiers as baseline models. Hidden Markov Model and CRF Model gave the best results for our problem. Comparatively, the HMM gave less accuracy than the CRF Model. The main reason for predicting the wrong language tag is the variation in tag used in the train data of English Telugu words. Our best performance system for tagging the language tag for a word is conditional random field with f1-score: 0.91 and accuracy: 91.2897\%.

\begin{table}[h!]
\begin{center}
\begin{tabular}{|c|c|}
\hline 
\textbf{Model}  &  \textbf{Accuracy(\%)} \\ \hline
Naïve Bayes Classifier & 77.37 \\
Random Forest Classifier & 77.34 \\
Hidden Markov Model & 85.15 \\
Conditional Random Field & 91.28 \\

\hline
\end{tabular}
\end{center}
\caption{\label{font-table}  Consolidated Results (Accuracy). }
\end{table}

In this work some interesting problems are encountered like Romanization of Telugu words, different types of syntax in social media text...etc. Since there is no standard way to transliterate the code mixed data and Romanization contributes a lot to the spelling errors in foreign words. For example, a single Telugu word can have the more than one spelling (Eg. ``avaru'', ``evaru'', ``aivaru'', ``yevaru''. Translation into English: ``who''). This posed a significant challenge for language identification. 

Similarly, In social media, chat conversation using SMS language ``you'' can be written as ``U'', ``Hai'' – ``Hi'', ``Good'' – ``gooooood''...etc. Such non standard usage is an issue for language identification.

The results are encouraging and future work can be focused on obtaining more social media corpus and using deep learning approaches such as LSTM in future studies. 

\section*{Acknowledgments}

The authors would like to thank the Twelfth International Conference on Natural Language Processing (ICON -2015) for code-mixed dataset. We immensely grateful to Nikhilesh Bhatnagar for constructive criticism of the manuscript.

\end{document}